\pdfoutput=1
\PassOptionsToPackage{table}{xcolor}
\documentclass[11pt]{article}

\usepackage{acl}

\usepackage{makecell} 

\usepackage{times}
\usepackage{latexsym}
\usepackage{graphicx} 
\usepackage{soul,color}
\usepackage{bbm}
\usepackage{amsmath,amsthm,mathtools}
\usepackage{booktabs}
\usepackage{multirow}
\usepackage{float}
\usepackage{stfloats}
\usepackage{flushend}
\usepackage{hyperref}
\usepackage{balance}

\renewcommand{\arraystretch}{1.1}

\usepackage{placeins}

\usepackage[normalem]{ulem} 

\usepackage{enumitem}
\setlist[itemize]{align=parleft,left=0pt..1em}

\newcommand{\fscorestar}[3]{#1 {\scriptsize($\pm$#2)}\textsuperscript{\scriptsize#3}}

\newif\ifcomment

\ifcomment
  \newcommand{\simret}[1]{\textcolor[rgb]{0.6,0,0.2}{Simret: #1}}
  
  \newcommand{\deleted}[1]{\textcolor[rgb]{0.8,0.8,0.8}{#1}}
\else
  \newcommand{\simret}[1]{}
  
  \newcommand{\deleted}[1]{}
\fi

\usepackage[T1]{fontenc}
\usepackage[utf8]{inputenc}

\usepackage{microtype}

\usepackage{inconsolata}

\title{Leveraging Variation Theory in Counterfactual Data Augmentation for Optimized Active Learning}

\author{Simret Araya Gebreegziabher \\  sgebreeg@nd.edu \\ University of Notre Dame 
        \And Kuangshi Ai \\ kai@nd.edu \\ University of Notre Dame
        \And Zheng Zhang \\ zzhang37@nd.edu \\ University of Notre Dame 
        \AND
         Elena L. Glassman$^*$  \\ glassman@seas.harvard.edu \\ Harvard University
        \And Toby Jia-Jun Li$^*$ \\ toby.j.li@nd.edu \\ University of Notre Dame}

\begin{document}
\maketitle
% \def\thefootnote{*}\footnotetext{Co-senior authors contributed equally.}\def\thefootnote{\arabic{footnote}}

%  TLDR; Inspired by Variation Theory we use neuro-symbolic patterns to guide the generation of counterfactual examples using an LLM. We then evaluate the generated counterfactuals in their ability to address the cold start problem in active learning.
\begin{abstract}
Active Learning (AL) allows models to learn interactively from user feedback. However, only annotating existing samples may hardly benefit the model's generalization. Moreover, AL commonly faces a cold start problem due to insufficient annotated data for effective sample selection. To address this, we introduce a counterfactual data augmentation approach inspired by Variation Theory, a theory of \textit{human concept learning} that emphasizes the essential features of a concept by focusing on what stays the same and what changes. We use a neuro-symbolic pipeline to  pinpoint key conceptual dimensions and use a large language model (LLM) to generate targeted variations along those dimensions. Through a text classification experiment, we show that our approach achieves significantly higher performance when there are fewer annotated data, showing its capability to address the cold start problem in AL. We also find that as the annotated training data gets larger, the impact of the generated data starts to diminish. This work demonstrates the value of incorporating human learning theories into the design and optimization of AL.
\end{abstract}

\def\thefootnote{*}\footnotetext{Co-senior authors contributed equally.}

\section{Introduction}

Active learning~(AL) allows users to provide focused annotations to integrate human preferences and domain knowledge into machine learning models~\cite{settles2009active}. It relies on a human's iterative annotations to build and refine model performance~\cite{budd2021survey}. As a result, the model's performance improvement with each annotation round depends on both the quality and quantity of annotated data. However, AL faces a cold start problem: in the early stages, when annotated data is limited, the model is often unstable and struggles to make informed decisions about which instances to query for labeling, which hinders its initial performance~\cite{yuan2020cold}. 
% Previous work showed that careful selection of examples to be annotated is instrumental to achieve optimal performance gain~\cite{beck2013reducing}.

\begin{figure}[t]
\centering
\includegraphics[width=0.5\textwidth]{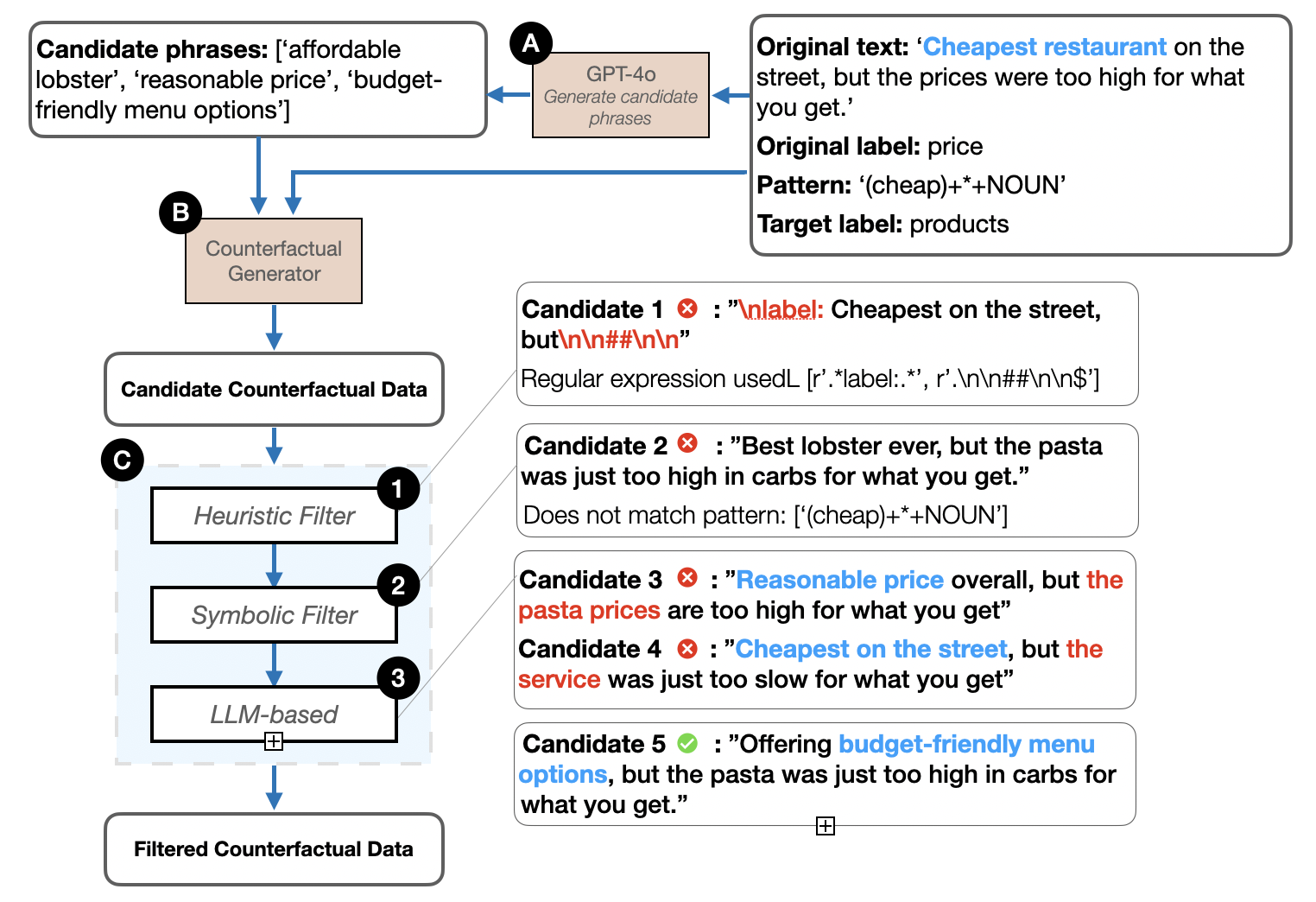}
\caption{Our approach combines neuro-symbolic patterns with in-context learning to generate Variation Theory-based counterfactual examples for active learning.}
\label{fig: our_pipeline}
\end{figure}

Counterfactual data augmentation techniques have been shown to enhance model performance~\cite{liu-etal-2021-counterfactual, yang2022exploring, wang2020robustness, reddy2023rethinking}. Synthesized counterfactual data can be more effective in capturing meaningful variations than real data selected from the dataset. However, the scalable generation and selection of augmented data have been a consistent challenge~\cite{liu-etal-2022-wanli, li2023large}. To address this, DISCO~\cite{chen2023disco} proposed a method for automatically generating counterfactual data using task-agnostic models. Despite its robust approach to augmented data, DISCO's use of a black-box pipeline makes debugging and improving the model difficult and does not allow meaningful presentation of variations that facilitate effective human annotation and sensemaking.

To address this, we propose a counterfactual generation pipeline that uses neuro-symbolic patterns to identify important features and uses them to guide the LLM's counterfactual generation\footnote{\href{https://github.com/SimretA/Variation-Theory-in-Counterfactual-Data-Augmentation}{www.github.com/SimretA/Variation-Theory-in-Counterfactual-Data-Augmentation}}. To motivate this approach, we draw on neuro-symbolic AI, which combines the representational power of neural networks with the interpretability and structure of symbolic reasoning~\cite{hitzler2022neuro}. Neuro-symbolic models integrate learned patterns with human-understandable rules, enabling systems to generalize in a transparent reasoning process. In the context of counterfactual generation, this hybrid approach allows us to generate examples that vary meaningfully along conceptually relevant dimensions while maintaining structural and semantic consistency. Specifically, we use a programming-by-example approach~\cite{gulwani2011automating} to generate neuro-symbolic patterns~\cite{PaTAT}. These patterns capture the syntactic and semantic similarities among similarly labeled examples. We then use the learned patterns to guide the LLM to generate counterfactual examples to be used in consecutive rounds of model re-training. The generated counterfactual examples change the assigned label into a different label while still keeping the original symbolic pattern in the data. In doing so, the generated examples introduce more meaningful variability in the data for subsequent model training. To further ensure the quality of the generated counterfactual examples, we design a three-step automatic filtering pipeline. 

This paper makes the following contributions: 

\paragraph{Evaluating the quality of generated counterfactual examples}We assess the quality of generated counterfactual examples using a three-stage filtering mechanism. We define a high-quality counterfactual as a sentence that eliminates the original label (soft flip) while introducing the target label (label flip). The results show a high Soft Label Flip Rate (SLFR)---the rate of removal of original labels from counterfactual examples, and a high level of consistency in Label Flip Rate (LFR)---the rate of changing original labels into target labels in generated counterfactual examples. By evaluating how often new examples meaningfully alter the original label and capture valuable variations, we can assess the efficacy of the examples produced. 

\paragraph{Evaluating the effectiveness of Variation Theory in active learning}We investigate how incorporating Variation Theory into active learning can improve robustness and address cold-start challenges~\cite{yuan2020cold}. Using a classification task, we compare our counterfactual-based method against four baselines—random, cluster-based, uncertainty-based selection, and counterfactuals without Variation Theory. Across three datasets and two models, our approach achieves up to 2× higher performance with fewer than 70 annotations. The benefits diminish as annotation volume grows, highlighting its effectiveness in low-data, cold-start settings. We also analyze the roles of annotation selection, syntactic diversity, and semantic diversity in driving this performance gain.

% This paper investigates the impacts of annotation selection, syntactic diversity, and semantic diversity of generated counterfactuals in active learning. We use a classification task to compare the performance of our method with four baseline conditions (random selection and cluster-based selection, uncertainty-based selection, and counterfactuals without Variation Theory). The results across three datasets and two models show that the use of counterfactual generated data results in at least two times higher performance with fewer number of annotations(<70) compared to the other conditions. As the number of annotated data increases, the impact of data augmentation starts to diminish, showing the efficacy of the approach in cold-start scenarios.

\looseness=-1

\section{Related Work}
\subsection{Active Learning}
Active Learning~(AL) in machine learning is an approach in which the learning algorithm selectively chooses informative data points for model training. Although most sampling strategies rely on a pool of unlabeled data~\cite{fu2013survey}, there are strategies that synthesize data points in real time for annotation~\cite{schumann2019active}. The second approach, also called Membership Query Synthesis~(MQS) creates new examples that inform the model with more representative scenarios by either modifying existing instances~\cite{wu2023scattershot, wu2021polyjuice} or generating new instances~\cite{schumann2019active}.

In domains with scarce annotated data, data augmentation methods aim to enhance the quantity and quality of training data~\cite{yang2022image}. Traditional data augmentation techniques, such as geometric transformations and color space alterations, do not modify the fundamental causal generative process. As a result, they do not counteract biases like spurious correlations~\cite{kaushik2021explaining}.

\subsection{Data Generation and Augmentation}

Counterfactual data augmentation has been widely used to counteract spurious correlations in data~\cite{denton2020image, liu-etal-2021-counterfactual, yang2022exploring, wang2020robustness}. This approach employs counterfactual inference to control generative factors, facilitating the generation of samples that can address confounding biases. Many existing strategies use dataset-specific counterfactual augmentation methods in specific domains, such as sentiment analysis~\cite{yang2022exploring, Kaushik2020Learning}, named entity recognition~\cite{Ghaddar2021Context}, text classification~\cite{wang2020robustness}, and neural machine translation~\cite{liu-etal-2021-counterfactual}. A popular approach to address spurious dependence in NLP datasets is to use human-guided counterfactual augmentation through crowd sourcing~\cite{kaushik2021explaining, joshi2022investigation}. This approach presents individuals with data and preliminary labels, asking them to modify the data for an alternate label while avoiding unnecessary edits~\cite{Kaushik2020Learning}. This method depends on human efforts and expertise to overcome the challenge of automatically translating raw text into important features.

LLMs have been shown to possess extensive generative capacity, making them useful tools for counterfactual data generation. \citet{li2023large} introduced a method utilizing LLMs to generate domain-specific counterfactual samples through prompt design, highlighting the alignment between the efficacy of LLMs in domain-specific counterfactual generation and their overall proficiency in that domain. Although in-context learning has been a promising direction to get LLMs to perform different tasks,  \citet{min-etal-2022-rethinking} identified several key factors that influence its effectiveness, including the demonstration of the label space, the input text distribution, and the overall sequence format.

A consistent challenge in counterfactual generation has been the scalable generation and selection of augmented data~\cite{liu-etal-2022-wanli, li2023large}. To address this, DISCO~\cite{chen2023disco} introduced a method for automatically generating high-quality counterfactual data using task-agnostic ``teacher and student'' models to allow classifier models to learn causal representations. DISCO uses a neural syntactic parser to select the spans of the sentence to vary on to generate data using Large Language Models (LLMs). Although DISCO provides more robust models trained on augmented data, the use of black-box approaches to generate data could make model debugging and improvement harder. To address this, we adopt a neuro-symbolic approach to define the concept boundaries in user annotations~\cite{PaTAT}.

\subsection{Example-based Learning via Variation Theory}
Based on previous studies on LLMs as counterfactual generators, our work seeks to learn from human cognition and example-based learning to better guide LLMs to generate higher quality data. \textit{Will educational theories that work for human learners also work for AI?} Decades of research have demonstrated that using example-based learning constitutes an effective instructional strategy for humans acquiring new skills~\cite{Gog2010example}. Few-shot learning is an example-based learning method commonly used by LLMs. 

How can we use human learning theories to support the annotation of data and training of LLM classifiers? Variation Theory~\cite{marton2014necessary}, rooted in human learning research, gives us insights from human experience, e.g.,~\cite{Cheng2016LearningTT}. The core concept of this theory involves presenting sets of examples that vary along specific dimensions, enabling learners to identify and conceptualize the dimensions as a useful coordinate space for describing instantiations of the underlying concept. This aligns with the foundational principle of counterfactual data augmentation in machine learning.

\section{Approach} 
Our approach applies the Variation Theory of human learning to machine learning in the context of active learning~(AL). We propose a new approach of counterfactual data generation by combining neuro-symbolic methods and LLMs. Specifically, we use domain-specific neuro-symbolic patterns to learn the syntactic representation of similarly labeled data that define a neuro-symbolic model's learning space and concept boundaries. We then use the learned patterns to guide the generation of augmented data that helps a classification model learn important nuances about each label~(Fig.~\ref{fig: our_pipeline}-A,B). 

Through this approach we generate counterfactual data that are \textit{syntactically similar} to their original counterparts but semantically belong to a different label. To ensure the quality of the generated counterfactuals, we apply a three-level filtering mechanism~(Fig.~\ref{fig: our_pipeline}-C). \looseness=-1

\subsection{Using Neuro-symbolic Patterns to Define Concept Space}
\label{sec:rule-based}
% \sgcomment{Refers to A and B in Fig 2}
{Variation Theory suggests that humans learn a concept most effectively when they are shown examples that vary in only one specific dimension at a time, while all other aspects stay the same. Therefore, an important aspect of Variation Theory is determining which features should vary to emphasize their effects in the learning process.} {We achieve this by learning critical features from labeled data by generating neuro-symbolic patterns and make small modifications on the original sentence while maintaining consistency along the generated pattern}.

\subsubsection{Learning Neuro-symbolic Patterns}
We use a programming-by-example~\cite{lieberman2001your} approach to establish the boundaries of concepts defined by data points and their associated ground truth labels. While our simulation study currently relies on ground truth labels, these will be substituted with human annotations in forthcoming interactive systems. After we randomly select a few annotations, we use PaTAT's~\cite{PaTAT} interactive program synthesis approach to generate domain-specific pattern rules that match the annotated examples. These pattern rules represent the lexical, syntactic, and semantic similarities of data under the same label. PaTAT's pattern language includes the following components:
\begin{itemize}
    \item Part-of-speech (POS) tags: \texttt{VERB}, \texttt{PROPN}, \texttt{NOUN}, \texttt{ADJ}, \texttt{ADV}, \texttt{AUX}, \texttt{PRON}, \texttt{NUM}
    \item Word stemming: \texttt{[WORD]} (e.g., \texttt{[have]} will match all variants of have, such as \textit{had}, \textit{has}, and \textit{having})
    \item Soft match: \texttt{(word)} (e.g., \texttt{(pricey)} will match synonyms such as \textit{expensive} and \textit{costly}, etc.)
    \item Entity type: \texttt{\$ENT-TYPE} (e.g., \texttt{\$LOCATION} will match phrases of location type, such as \textit{Houston, TX} and \textit{California}; \texttt{\$DATE} will match dates; \texttt{\$ORG} will match names of organizations)
    \item Wildcard: \texttt{*} (will match any sequence of words)
\end{itemize}
\noindent Although the fundamental patterns are suitable for general domain text data, it is feasible to expand the pattern language to include specialized or domain-specific patterns.

This method generates a collection of regex-like patterns~(but with semantically-enhanced tags) that match with the labeled positive examples while excluding the labeled negative examples.
For example, if two data points in the domain of restaurant review \textit{``Good food with great variety."} and \textit{``The food was amazing."} have the same label ``products", PaTAT learns up to 5 patterns that collectively match the set of examples annotated with that label. In this case, two patterns match both sentences, i.e., \textit{``[food]+*+ADJ}'', \textit{``(amazing)+*''}.

\subsubsection{Using Neuro-symbolic Patterns for Counterfactual Data Generation}
% \sgcomment{Refers to A and B in Fig 2}

Using the learned neuro-symbolic patterns, we generate counterfactual examples by modifying the original text to be about a different label while still keeping the original pattern. To ensure minimal modifications and to make sure the reason for the original label is kept, we begin by generating candidate phrases for segments of the original sentence that matched the neuro-symbolic pattern~(Fig.~\ref{fig: our_pipeline}-A). 

We use the generated candidate phrases as constraints to be included in the generated sentence. For example, in Fig.~\ref{fig: our_pipeline}, the pattern \textit{(cheap)+*+NOUN} has candidate phrases \textit{[`affordable lobster', `reasonable price', `budget-friendly menu']}. When generating the counterfactual example, we instruct the LLM to always include one of these phrases in the modified sentence. This constraint ensures that counterfactual examples that vary in semantic content remain within the syntactic boundaries set by the pattern, which defines, at least in part, the particular label for which counterexamples are being generated~(Fig.~\ref{fig: our_pipeline}-B).

\subsection{Filtering Generated Counterfactual Data}
\label{sec: filter}
% \sgcomment{Refers to 3 - A, B, C in Fig 2}
The ideal counterfactual example is a complete and coherent sentence that should keep the patterns of the original text, and successfully flip the original label to the target label. To ensure the quality of the fine-tuning dataset, we implement a three-stage filtering mechanism:

\subsubsection{Regex Heuristic Filtering}
We use a heuristic-based filter to identify and remove counterfactual data with common generation flaws. This filter ensures that the generated sentences are coherent and complete. This method uses regular expressions to detect common generation errors observed during our experimentation~(Fig.~\ref{fig: our_pipeline}-C1). We define rules to identify error patterns such as repetition of the prompt, inaccurate formatting, and incomplete generation, which were some common pitfalls we observed during generation.

\subsubsection{Neuro-symbolic Filtering}
The neuro-symbolic filter ensures that the generated counterfactual examples retain the original learned pattern. The original patterns represent features the model learns as useful conceptual boundaries. Therefore, keeping them in the counterfactually generated examples challenges the model's current boundary. To achieve this, we implement the filter using executable neuro-symbolic patterns defined in \S~\ref{sec:rule-based}. Specifically, we check whether each generated counterfactual example matches its original counterpart's neuro-symbolic pattern~(Fig.~\ref{fig: our_pipeline}-C2). This filter excludes generated counterfactual examples that do not match with the provided pattern from being used in the consecutive training pipeline. 
To quantify this over the generated counterfactual examples, we calculate the pattern keeping rate~(PKR) as defined below. 
\begin{gather*}
PKR=\frac{1}{N}\sum _{n=1}^{N}\mathbbm{1}(\hat{p}_{n} =p_{n})
\end{gather*}
where $\displaystyle p_{n}$ is the original pattern, $\displaystyle \hat{p}_{n}$ is the pattern given to the counterfactual data, and $\displaystyle {N}$ is the size of the counterfactual data.

\subsubsection{LLM-based Discriminator Filtering}
Finally, we apply a filter using a GPT-4o discriminator. This filter removes counterfactuals that still keep their original label and all counterfactuals that do not change the label to the target label~(Fig.~\ref{fig: our_pipeline}-C3). This filter makes sure that the generated counterfactual examples have enough semantic changes that changes the original label to the target label. We adopt two matrices~\cite{chen2023disco} to quantify this: the Label Flip Rate~(LFR), and the Soft Label Flip Rate~(SLFR) as defined below:

\begin{gather*}
LFR=\frac{1}{N}\sum _{n=1}^{N}\mathbbm{1}\left(\hat{l}_{n} ={L}_{n}\right)\\
SLFR=\frac{1}{N}\sum _{n=1}^{N}\mathbbm{1}(\hat{l}_{n} \neq l_{n})\\
\end{gather*}
where $\displaystyle \hat{l}_{n}$ is the label given by GPT-4o discriminator, $\displaystyle {L}_{n}$ is the target label, $\displaystyle l_{n}$ is the original label.

SLFR measures the rate at which the generated counterfactual remove their original label, and LFR evaluates how often the counterfactual examples successfully adopt the target label.

\section{Experiments}
We evaluate the generated counterfactuals using two experiments\footnote{We spent approximately 400USD in total on API calls to OpenAI for running Experiments 1 and 2. Since the running of the experiments, the cost of GPT-4o has decreased by 79\%.}. First, we evaluate the quality of generated counterfactual examples using the PKP, LFR, and SLFR metrics in \S~\ref{sec: filter}.

In the second experiment, we compare our proposed approach to other example selection techniques in a standard classification task, using two pre-trained models. We use five different data selection techniques in interactive AL: random selection, cluster-based selection, uncertainty-based selection, counterfactual examples generated without Variation Theory, and our proposed counterfactual based example selection. We use each dataset's original label as ground truth and use GPT-4o and a BERT model as the target classification models. 

To further understand the impact of each component of our filtering pipeline, we conduct an ablation study. In this study, we aim to understand the impact of each individual filter on the pipeline's performance in downstream model training. Additional details can be found in Appendix \ref{sec:ablation study}.

\subsection{Datasets}

\begin{itemize}
\item \textbf{YELP}: The YELP dataset~\cite{asghar2016yelp} consists of user reviews of different businesses and services. The dataset itself provides 4 ground-truth categories (i.e. service, price, environment and products), we randomly sampled 495 examples for this experiment.
\item \textbf{MASSIVE}: The MASSIVE~\cite{fitzgerald2022massive} virtual assistant utterances with 18 labeled intents as ground-truth (e.g. audio, cooking, weather, recommendation etc). For this experiment we randomly selected 30 examples from each category, making up a total of 540 examples.

\item \textbf{Emotions}: Includes a collection of English Twitter messages annotates with 6 emotions: anger, fear, joy, love, sadness, and surprise~\cite{elgiriyewithana_emotions_2024}. For this experiment we randomly selected 500 examples while balancing the number of labels.

\end{itemize}

\subsection{Experiment 1: Generated Counterfactual Quality}

We evaluate the generated counterfactuals using two experiments. First, we evaluate the quality of generated counterfactual examples using the PKP, LFR, and SLFR metrics in \S~\ref{sec: filter}.

\subsubsection{Results}

\begin{table}[H]
\centering
\small
\begin{tabular}{rccc}
\toprule
\rowcolor{blue!5}
 & YELP & MASSIVE & Emotions \\ \midrule
Pattern Keeping Rate & 0.94 & 0.88 & 0.81 \\
Soft Label Flip Rate & 0.45 & 0.71 & 0.58 \\
Label Flip Rate & 0.98 & 0.86 & 0.86 \\
\bottomrule
\end{tabular}
\caption{Generated counterfactual data quality evaluation.}
\label{tab:counter_quality}
\end{table}
Our findings indicate that our proposed pipeline maintains the quality of generated counterexamples, as measured by Pattern Keeping Rate (PKR) and Label Flip Rate (LFR). Across datasets, the PKR remains high, demonstrating the generated counterfactual examples effectively keep the original pattern rules. The LLM-based Discriminator Filtering achieves robust performance in LFR across datasets, confirming that most counterfactual examples successfully adopt the target label. However, the Soft Label Flip Rate (SLFR) varies, particularly with the MASSIVE dataset showing the highest rate and the others on the lower side. This suggests that the degree of semantic change required to remove the original label can be dataset-dependent.

\subsection{Experiment 2: Generated Counterfactuals in Downstream Model Training}
\label{sec:conditions}

In the second experiment, we compare our counterfactual generation approach with five other sampling strategies in AL.

\begin{itemize}
\item \textbf{Random} Examples are randomly selected for each annotation iteration to train the classification model.

\item \textbf{Cluster} Examples selected from a k-means clustered, pretrained Sentence Transformer model by iterating through the clusters in rotation.

\item \textbf{Uncertainty} We use model confidence on the training set to choose data with the lowest confidence to be labeled. We use verbal uncertainty~\cite{lin2022teaching} to get model confidence in GPT-4o and model logits for the BERT model.

\item \textbf{ALPS~\cite{yuan2020cold}} We use ALPS a sampling strategy that addresses the cold start problem in AL.

\item \textbf{Counterexamples without Variation Theory} We generate counterexamples without using the neuro-symbolic pipeline defined in Fig~\ref{fig: our_pipeline}.

\end{itemize}

\subsubsection{Protocol}
\label{sec:active_learning_eval}

\begin{table*}[tbp!]
\centering
\scriptsize
\setlength{\tabcolsep}{8pt}
\renewcommand{\arraystretch}{1.0}

\textbf{Macro F1-scores (GPT-4o)} \\

\vspace{.3em}

% YELP - GPT-4o
\scriptsize{YELP} \\

\resizebox{\textwidth}{!}{\begin{tabular}{rlllllll}
\toprule
\rowcolor{blue!5}
\textbf{Method} & 10 & 15 & 30 & 50 & 70 & 90 & 120 \\
\midrule
Random & \fscorestar{.38}{.05}{***} & \fscorestar{.44}{.06}{***} & \fscorestar{.51}{.07}{***} & \fscorestar{.61}{.05}{} & \fscorestar{.65}{.06}{} & \fscorestar{.69}{.04}{+} & \fscorestar{.74}{.04}{} \\
Cluster & \fscorestar{.41}{.07}{***} & \fscorestar{.48}{.04}{***} & \fscorestar{.57}{.07}{} & \fscorestar{.63}{.06}{} & \fscorestar{.68}{.03}{*} & \fscorestar{.69}{.03}{+} & \fscorestar{.70}{.02}{} \\
Uncertainty & \fscorestar{.23}{.04}{***} & \fscorestar{.21}{.05}{***} & \fscorestar{.27}{.06}{***} & \fscorestar{.28}{.05}{***} & \fscorestar{.29}{.04}{***} & \fscorestar{.28}{.06}{***} & \fscorestar{.29}{.05}{} \\
ALPS & \fscorestar{.37}{.04}{**} & \fscorestar{.49}{.06}{*} & \cellcolor{green!20} \fscorestar{\textbf{.66}}{.05}{} & \fscorestar{.68}{.03}{} & \cellcolor{green!20} \fscorestar{\textbf{.69}}{.03}{} & \cellcolor{green!20} \fscorestar{\textbf{.70}}{.04}{} & \fscorestar{.72}{.03}{} \\
Counterfactuals without VT &  \fscorestar{.35}{.10}{***} & \fscorestar{.46}{.13}{*} & \fscorestar{.54}{.05}{*} & \fscorestar{.53}{.06}{*} & \fscorestar{.39}{.08}{***} & \fscorestar{.25}{.05}{***} & \fscorestar{.31}{.05}{} \\
Counterfactuals & \cellcolor{green!20} \fscorestar{\textbf{.55}}{.08}{} & \cellcolor{green!20} \fscorestar{\textbf{.59}}{.07}{} &\fscorestar{.63}{.07}{} & \cellcolor{green!20} \fscorestar{\textbf{.69}}{.07}{} & \fscorestar{.59}{.10}{} & \fscorestar{.65}{.05}{} & \cellcolor{green!20} \fscorestar{\textbf{.78}}{.04}{} \\

\bottomrule
\end{tabular}}

\vspace{.2em}

% MASSIVE - GPT-4o
\scriptsize{MASSIVE} \\
\resizebox{\textwidth}{!}{
\begin{tabular}{rlllllll}
\toprule
\rowcolor{blue!5}
\textbf{Method} & 10 & 15 & 30 & 50 & 70 & 90 & 120 \\
\midrule
Random &  \fscorestar{.36}{.06}{***} & \fscorestar{.40}{.05}{*} & \fscorestar{.49}{.12}{} & \fscorestar{.51}{.11}{} & \fscorestar{.54}{.10}{*} & \fscorestar{.57}{.09}{***} & \fscorestar{.61}{.10}{} \\
Cluster & \fscorestar{.35}{.06}{***} & \fscorestar{.40}{.07}{*} & \fscorestar{.47}{.08}{} & \fscorestar{.49}{.08}{} & \fscorestar{.56}{.12}{*} & \fscorestar{.54}{.12}{*} & \fscorestar{.55}{.09}{} \\
Uncertainty & \fscorestar{.22}{.08}{***} & \fscorestar{.19}{.10}{***} & \fscorestar{.18}{.07}{***} & \fscorestar{.13}{.06}{***} & \fscorestar{.14}{.07}{***}& \fscorestar{.19}{.09}{***} & \fscorestar{.20}{.10}{} \\
ALPS & \fscorestar{.12}{.03}{*} & \fscorestar{.24}{.08}{} & \fscorestar{.39}{.02}{} & \cellcolor{green!20} \fscorestar{\textbf{.61}}{.03}{} & \cellcolor{green!20} \fscorestar{\textbf{.65}}{.08}{} & \cellcolor{green!20} \fscorestar{\textbf{.67}}{.07}{} & \fscorestar{.72}{.04}{} \\
Counterfactuals without VT & \fscorestar{.26}{.10}{***} & \fscorestar{.37}{.07}{*} & \fscorestar{.43}{.05}{*} & \fscorestar{.40}{.07}{} & \fscorestar{.34}{.10}{} & \fscorestar{.27}{.09}{*} & \fscorestar{.37}{.08}{} \\
Counterfactuals & \cellcolor{green!20} \fscorestar{\textbf{.48}}{.01}{} & \cellcolor{green!20} \fscorestar{\textbf{.52}}{.03}{} & \cellcolor{green!20} \fscorestar{\textbf{.59}}{.3}{} & \fscorestar{.63}{.03}{} & \fscorestar{.64}{.06}{} & \fscorestar{.66}{.05}{} & \cellcolor{green!20} \fscorestar{\textbf{.79}}{.03}{} \\
\bottomrule
\end{tabular}
}
\vspace{.2em}

% EMOTIONS - GPT-4o
\scriptsize{EMOTIONS} \\
\resizebox{\textwidth}{!}{\begin{tabular}{rlllllll}
\toprule
\rowcolor{blue!5}
\textbf{Method} & 10 & 15 & 30 & 50 & 70 & 90 & 120 \\
\midrule
Random & \fscorestar{.29}{.10}{} & \fscorestar{.32}{.10}{} & \fscorestar{.36}{.07}{***} & \fscorestar{.39}{.04}{***} & \fscorestar{.45}{.04}{*} & \fscorestar{.45}{.06}{} & \fscorestar{.47}{.04}{} \\
Cluster & \fscorestar{.32}{.04}{} & \fscorestar{.38}{.04}{} & \fscorestar{.36}{.08}{***} & \fscorestar{.39}{.12}{***} & \fscorestar{.42}{.09}{*} & \fscorestar{.42}{.08}{} & \fscorestar{.41}{.05}{} \\
Uncertainty & \fscorestar{.21}{.07}{***} & \fscorestar{.19}{.05}{***} & \fscorestar{.25}{.05}{***} & \fscorestar{.29}{.04}{***} & \fscorestar{.28}{.07}{***} & \fscorestar{.29}{.06}{} & \fscorestar{.33}{.05}{} \\
ALPS & \fscorestar{.23}{.07}{} & \fscorestar{.26}{.03}{} & \fscorestar{.34}{.05}{} & \fscorestar{.36}{.05}{} & \fscorestar{.39}{.06}{} & \fscorestar{.40}{.05}{} & \fscorestar{.44}{.10}{} \\
Counterfactuals without VT & \fscorestar{.28}{.06}{} & \fscorestar{.35}{.10}{} & \fscorestar{.46}{.13}{} & \fscorestar{.48}{.13}{} & \fscorestar{.49}{.12}{} & \fscorestar{.36}{.08}{} & \fscorestar{.39}{.07}{} \\
Counterfactuals & \cellcolor{green!20} \fscorestar{\textbf{.34}}{.08}{} & \cellcolor{green!20} \fscorestar{\textbf{.43}}{.10}{} & \cellcolor{green!20} \fscorestar{\textbf{.54}}{.10}{} & \cellcolor{green!20} \fscorestar{\textbf{.51}}{.05}{} & \cellcolor{green!20} \fscorestar{\textbf{.58}}{.10}{} & \cellcolor{green!20} \fscorestar{\textbf{.47}}{.03}{} & \cellcolor{green!20} \fscorestar{\textbf{.52}}{.05}{} \\
\bottomrule
\end{tabular}
}

\caption{Macro F1-scores for GPT-4o across three datasets (YELP, MASSIVE, EMOTIONS) with varying annotation shot counts. + indicates p-value<.1, * indicates p-value<.05, ** indicates p-value<.01, and *** shows p-value<.0001 between the condition and the counterfactual condition.}
\label{tab:gpt4o-scores}
\end{table*}

\begin{table*}[tbp]
\centering
\small
\setlength{\tabcolsep}{4pt}
\renewcommand{\arraystretch}{1.2}

\textbf{Macro F1-scores (BERT)} \\

\vspace{.3em}

% YELP - BERT
\scriptsize{YELP} \\
\resizebox{\textwidth}{!}{\begin{tabular}{rlllllllll}
\toprule
\rowcolor{blue!5}
\textbf{Method} & 10 & 15 & 30 & 50 & 70 & 90 & 120 & 150 & 170 \\
\midrule
Random & \fscorestar{.16}{.06}{*} & \fscorestar{.18}{.05}{***} & \fscorestar{.26}{.03}{***} & \fscorestar{.33}{.04}{***} & \fscorestar{.35}{.06}{***} & \fscorestar{.45}{.01}{} & \fscorestar{.45}{.03}{} & \fscorestar{.48}{.04}{} & \fscorestar{.51}{.02}{} \\
Cluster & \fscorestar{.18}{.08}{***} & \fscorestar{.19}{.06}{***} & \fscorestar{.26}{.07}{***} & \fscorestar{.32}{.06}{***} & \fscorestar{.34}{.05}{+} & \fscorestar{.46}{.03}{} & \fscorestar{.31}{.08}{} & \fscorestar{.42}{.1}{} & \fscorestar{.45}{.1}{} \\
Uncertainty & \fscorestar{.13}{.06}{} & \fscorestar{.14}{.04}{} & \fscorestar{.19}{.07}{} & \fscorestar{.33}{.04}{} & \fscorestar{.41}{.06}{} & \fscorestar{.46}{.03}{} & \fscorestar{.47}{.04}{} & \fscorestar{.53}{.04}{} & \fscorestar{.54}{.03}{} \\
ALPS & \fscorestar{.14}{.05}{} & \fscorestar{.16}{.06}{} & \fscorestar{.15}{.06}{} & \fscorestar{.25}{.08}{} & \fscorestar{.27}{.08}{} & \fscorestar{.27}{.08}{} & \fscorestar{.36}{.11}{} & \fscorestar{.37}{.11}{} & \fscorestar{.37}{.10}{} \\
Counterfactuals without VT & \fscorestar{.20}{.06}{} & \fscorestar{.16}{.07}{} & \fscorestar{.25}{.04}{} & \fscorestar{.29}{.04}{} & \fscorestar{.38}{.08}{} & \fscorestar{.45}{.05}{} & \fscorestar{.49}{.04}{} & \cellcolor{green!20}  \fscorestar{\textbf{.54}}{.05}{} & \cellcolor{green!20}  \fscorestar{\textbf{.55}}{.04}{} \\
Counterfactuals & \cellcolor{green!20}  \fscorestar{\textbf{.38}}{.04}{} & \cellcolor{green!20} 
 \fscorestar{\textbf{.39}}{.07}{} & \cellcolor{green!20}  \fscorestar{\textbf{.49}}{.05}{} & \cellcolor{green!20} \fscorestar{\textbf{.47}}{.04}{} & \cellcolor{green!20} 
 \fscorestar{\textbf{.51}}{.04}{} & \cellcolor{green!20}  \fscorestar{\textbf{.53}}{.04}{} & \cellcolor{green!20}  \fscorestar{\textbf{.50}}{.03}{} & \fscorestar{.52}{.02}{} & \fscorestar{.53}{.03}{} \\
\bottomrule
\end{tabular}
}

\vspace{.2em}

% MASSIVE - BERT
\scriptsize{MASSIVE} \\
\resizebox{\textwidth}{!}{\begin{tabular}{rlllllllll}
\toprule
\rowcolor{blue!5}
\textbf{Method} & 10 & 20 & 30 & 50 & 70 & 100 & 130 & 150 & 170 \\
\midrule
Random & \fscorestar{.048}{.03}{***} & \fscorestar{.052}{.03}{***} & \fscorestar{.12}{.04}{***} & \fscorestar{.11}{.05}{***} & \fscorestar{.19}{.03}{***} & \fscorestar{.22}{.02}{***} & \fscorestar{.23}{.02}{***} & \fscorestar{.24}{.02}{***} & \fscorestar{1}{.02}{} \\

Cluster & \fscorestar{.046}{.01}{***} & \fscorestar{.058}{.04}{***} & \fscorestar{.091}{.03}{***} & \fscorestar{.13}{.04}{***} & \fscorestar{.18}{.04}{***} & \fscorestar{.20}{.03}{***} & \fscorestar{.23}{.02}{***} & \fscorestar{.24}{.02}{***} & \fscorestar{.25}{.02}{} \\

Uncertainty & \fscorestar{.029}{.02}{***} & \fscorestar{.035}{.02}{***} & \fscorestar{.11}{.04}{***} & \fscorestar{.14}{.03}{***} & \fscorestar{.22}{.02}{***} & \fscorestar{.23}{.03}{***} & \fscorestar{.24}{.03}{***} & \fscorestar{.25}{.03}{***} & \fscorestar{.25}{.02}{***} \\

ALPS & \fscorestar{.017}{.01}{***} & \fscorestar{.13}{.01}{***} & \fscorestar{.14}{.01}{***} & \fscorestar{.19}{.01}{***} & \fscorestar{.31}{.01}{} & \fscorestar{.23}{.01}{} & \fscorestar{.45}{.02}{} & \fscorestar{.45}{.02}{} & \fscorestar{.64}{.05}{} \\

Counterfactuals without VT & \fscorestar{.09}{.08}{***} & \fscorestar{.15}{.07}{***} & \fscorestar{.33}{.08}{***} & \fscorestar{.50}{.07}{*} & \cellcolor{green!20}  \fscorestar{\textbf{.61}}{.05}{+} & \cellcolor{green!20} \fscorestar{\textbf{.64}}{.04}{} & \cellcolor{green!20} \fscorestar{\textbf{.68}}{.04}{*} & \cellcolor{green!20}  \fscorestar{\textbf{.68}}{.04}{} & \cellcolor{green!20} \fscorestar{\textbf{.69}}{.03}{+} \\

Counterfactuals &  \cellcolor{green!20}  \fscorestar{\textbf{.33}}{.09}{} & \cellcolor{green!20} 
 \fscorestar{\textbf{.40}}{.07}{} & \cellcolor{green!20}  \fscorestar{\textbf{.51}}{.08}{} & \cellcolor{green!20}  \fscorestar{\textbf{.58}}{.06}{} & \fscorestar{.56}{.05}{} & \fscorestar{.60}{.09}{} & \fscorestar{.61}{.06}{} & \fscorestar{.66}{.05}{} & \fscorestar{.62}{.1}{} \\
\bottomrule
\end{tabular}
}

\vspace{.2em}

% EMOTIONS - BERT
\scriptsize{EMOTIONS} \\
\resizebox{\textwidth}{!}{\begin{tabular}{rlllllllll}
\toprule
\rowcolor{blue!5}
\textbf{Method} & 10 & 20 & 30 & 50 & 70 & 100 & 130 & 150 & 170 \\
\midrule

Random &\fscorestar{.19}{.04}{*} & \fscorestar{.20}{.03}{***} & \fscorestar{.24}{.08}{*} & \fscorestar{.31}{.12}{} & \fscorestar{.46}{.09}{} & \fscorestar{.47}{.09}{} & \fscorestar{.53}{.14}{} & \fscorestar{.63}{.07}{} & \fscorestar{.30}{.06}{} \\

Cluster &\fscorestar{.18}{.02}{*} & \fscorestar{.21}{.03}{*} & \fscorestar{.23}{.02}{***} & \fscorestar{.28}{.03}{*} & \fscorestar{.41}{.05}{} & \fscorestar{.43}{.08}{} & \fscorestar{.48}{.06}{} & \fscorestar{.59}{.05}{} & \fscorestar{.52}{.12}{} \\

Uncertainty & \fscorestar{.23}{.04}{***} & \fscorestar{.23}{.05}{} & \fscorestar{.26}{.08}{*} & \fscorestar{.35}{.05}{} & \fscorestar{.38}{.04}{+} & \cellcolor{green!20}  \fscorestar{\textbf{.57}}{.07}{***} & \cellcolor{green!20}  \fscorestar{\textbf{.66}}{.08}{***} & \fscorestar{.69}{.07}{} & \fscorestar{.70}{.06}{*} \\

ALPS & \fscorestar{.09}{.04}{} & \fscorestar{.15}{.04}{} & \fscorestar{.28}{.04}{}& \fscorestar{.24}{.04}{} & \fscorestar{.42}{.04}{}& \fscorestar{.44}{.03}{} & \fscorestar{.52}{.03}{} & \cellcolor{green!20} 
 \fscorestar{\textbf{.74}}{.03}{} &\cellcolor{green!20}  \fscorestar{\textbf{.75}}{.03}{}  \\

Counterfactuals without VT & \fscorestar{.18}{.05}{*} & \fscorestar{.21}{.05}{*} & \fscorestar{.32}{.09}{} & \fscorestar{.36}{.12}{} & \fscorestar{.40}{.13}{} & \fscorestar{.57}{.08}{***} & \fscorestar{.62}{.1}{} & \fscorestar{.62}{.2}{} & \fscorestar{.72}{.05}{*} \\

Counterfactuals & \cellcolor{green!20}  \fscorestar{\textbf{.27}}{.07}{} & \cellcolor{green!20} 
 \fscorestar{\textbf{.26}}{.09}{} & \cellcolor{green!20}  \fscorestar{\textbf{.36}}{.05}{} & \cellcolor{green!20}  \fscorestar{\textbf{.38}}{.12}{} & \cellcolor{green!20}  \fscorestar{\textbf{.49}}{.05}{} & \fscorestar{.45}{.15}{} & \fscorestar{.50}{.06}{} & \fscorestar{.63}{.06}{} & \fscorestar{.56}{.07}{} \\

\bottomrule
\end{tabular}}

\caption{Macro F1-scores for BERT model across three datasets (YELP, MASSIVE, EMOTIONS) with varying annotation shot counts. + indicates p-value<.1, * indicates p-value<.05, ** indicates p-value<.01, and *** shows p-value<.0001 between the condition and the counterfactual condition.}
\label{tab:bert-scores}
\end{table*}

To evaluate the generated counterfactual examples, we employ a simulated active learning task to train and evaluate a BERT model~\cite{devlin2018bert} and few-shot prompting GPT-4o model for a multi-class classification task. We use the example selection conditions defined in \S~\ref{sec:conditions} to define a subset of 10, 15, 30, and progressively increasing up to 170 data points (referred to as `shots'), alongside their corresponding ground truths to be used as training sets. We then evaluate the classifier model using a hold-off set of the dataset.

To augment the model's training with generated counterfactual examples, we pair each original data with its generated counterfactual examples and their assigned target label. This pairing is used to enrich the distribution and quality of the training data, hypothesizing that the inclusion of counterfactuals would enhance the model's learning and predictive accuracy in early stages of annotation, addressing the cold start problem~\cite{yuan2020cold}. Similarly, the performance of the model, in this case trained with both original and counterfactual datasets, was again evaluated against the same hold-off set. This comparative analysis aimed to quantify the impact of counterfactual examples on the model's ability to generalize and make accurate predictions on unseen data in early active learning scenarios.

\subsubsection{Results}

We present our findings on the efficacy of generated counterfactuals in active learning as defined in \S~\ref{sec:active_learning_eval}. We report the macro F1-scores for the three datasets across different shots and conditions~(Table~\ref{tab:gpt4o-scores} and Table~\ref{tab:bert-scores}) using two models - few-shot learning with GPT-4o and fine-tuning a BERT model. We use training shots ranging from 10 to 120 shots for GPT-4o to stay within OpenAI's token limit and 10 to 170 for the BERT model.

We conducted a pair-wise t-test between the counterfactual condition and the other baseline conditions to understand the impact of the proposed approach. The results across the three datasets highlight the strong initial impact that the counterfactual condition has in addressing the cold start problem. We consistently observe a statistically significant advantage of the counterfactual condition in lower shot numbers. As the number of annotated examples increases (50 shots and above in most cases), the difference in average F1-score decreases, suggesting the advantage of the counterfactual condition diminishes when more data become available. Similarly, we observe significant impacts of the counterfactual condition when using a few-shot approach with the GPT-4o~(Table~\ref{tab:gpt4o-scores}). However, we did not find results that consistently indicated a substantial difference between the random, cluster, and counterfactual without variation theory conditions after 50 shots of examples have been labeled. The results demonstrated the \textbf{performance advantage} of our proposed neuro-symbolic variation theory-based counterfactual data augmentation approach in cold-start scenarios for active learning tasks.

Our approach introduces useful data to address the lack of label distribution and representation in cold start scenarios. Compared to the \textit{counterfactuals without Variation Theory} condition, the counterexamples generated through Variation Theory have a significantly higher F1-score, showing the impact of the pipeline in generating useful data in early AL. Moreover, the ablation study in Appendix~\ref{sec:ablation study} evaluating the impact of the filtering components in the pipeline shows there is a statistically significant difference in the downstream performance of a model trained on filtered data compared to data that does not have the complete filtering pipeline. 

As we get more annotated data, we observe either minimal improvement or a decline in the model's performance. We believe that this occurs because after a certain point, the generated counterfactuals begin to replicate previously observed patterns, and there is a limit to the amount of information that can be extracted from these patterns. We also see similar patterns of model decline in the \textit{counterfactuals without VT condition}. This ultimately may cause the model to overly rely on itself, resulting in the performance not scaling. To address this, it is important to heuristically understand the amount of data distribution that can be captured by generated data and switch gears back to using real data when needed. 

\section{Conclusion}
\citet{li2023synthetic} find that the performance of synthetic data is highly dependent on the distribution of the generated data, suggesting that enhancing data diversity could significantly improve the utility of synthetic data. Our approach achieves this by generating counterfactual examples along dynamic neuro-symbolic boundaries to allow the synthetic data to represent underlying concepts for better generalizability. In our evaluation, we find that models trained on counterfactual examples have a statistically significant advantage in the early stage of active learning, where there is a limited number of annotated data. When there is only a small amount of annotated data available, the distribution of the \textit{`real data'} may not sufficiently cover the latent space. 

Notably, the performance benefit of the counterfactual condition begins to decline when more than 70 labeled data points are used in model training. This reduction in advantage could potentially be attributed to model collapse. This happens when the model fails to capture the full diversity of the data on which it is trained~\cite{wang2023robustness, su2023beware}. With the introduced distribution shift, after the 70-shot threshold, the model might overfit to the specific characteristics of the synthetic examples it has seen, rather than generalizing to the broader real data distribution.

In terms of cost, we spent approximately 400USD in total on API calls to OpenAI for running Experiments 1 and 2. Since then, the cost of GPT-4o has decreased by 79\%, and we anticipate that the cost of using LLMs will continue to decline as the technology advances. Recent models are already demonstrating state-of-the-art performance at significantly lower costs. To explore the feasibility of more scalable alternatives, we also conducted an experiment using an open-weight model and found the results to be comparable (see Appendix~\ref{sec:llama-experiment}).
\section{Limitations}
% \hl{Separate limitation and future work. }
Our neuro-symbolic pipeline enables the automatic, real-time creation of counterfactual data using a pattern-based program synthesis approach. This method defines the concept space varied during counterfactual generation. Although the current pattern building blocks are designed for general domains, they rely on predefined rules, which may need augmentation with domain-specific lexical rules for specialized applications. Additionally, our use of a GPT-based discriminator to assign target labels for each counterfactual introduces potential biases or limitations inherent to the discriminator model itself. Future work could focus on understanding how human annotators understand and label the generated counterfactual examples.

\balance

% \balance

% \balance
\bibliography{custom}

\appendix
\section{Appendix}
\label{sec:appendix}

\subsection{Generation Pipeline}
\label{sec:appendix pipeline}
In this section, we provide the details of all the prompts and models we use to construct the whole counterfactual generation pipeline.

\subsubsection{GPT-4o Multi-label Separator}
\label{app:multilabel_separator}
As shown in Fig.~\ref{fig: prompt} Step-1, we utilize zero-shot GPT-4 to preprocess the raw data in order to separate the given multi-labeled sentences into several single-labeled parts. We call GPT-4 through the API provided by OpenAI, set the temperature parameter to 0, and restrict the maximum token number to 512, which ensures the reliability of the generated results. The prompt used is shown below:

\begin{figure*}[b]
\centering
\begin{minipage}{0.45\linewidth}
    \vspace{3pt}		\centerline{\includegraphics[width=\textwidth]{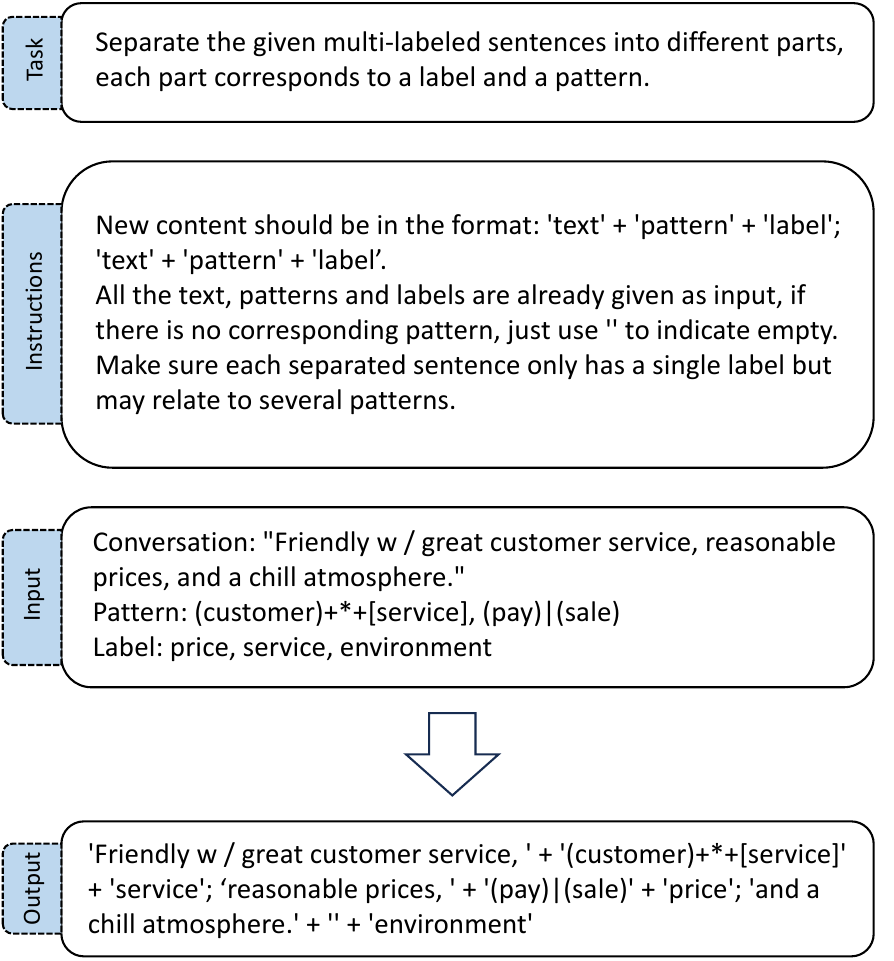}}
    \centerline{Step 1: separate multi-labeled text}
\end{minipage}
\begin{minipage}{0.45\linewidth}
    \vspace{3pt}
    \centerline{\includegraphics[width=\textwidth]{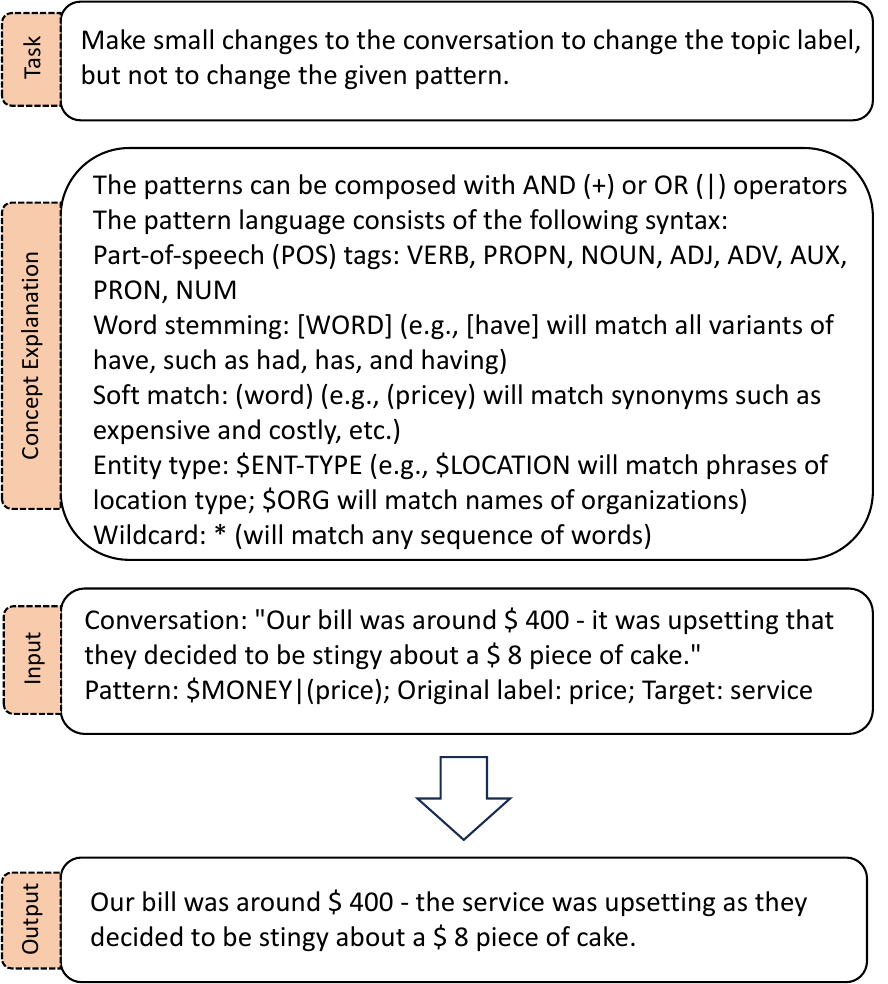}}
 
    \centerline{Step 2: generate pattern-kept counterfactual text}
\end{minipage}
\caption{Illustration of LLM prompts used for preparing training datapoints and generating counterfactual datapoints}
 \label{fig: prompt}
\end{figure*}
\raggedbottom

\small\begin{itemize}
\item \{``role'': ``system'', ``content'': ``The assistant will separate the given multi-labeled sentences into different parts, each corresponds to a label and a pattern (if the pattern is viable)''\}

\item \{``role": ``user", ``content": ``The assistant will generate outputs based on the following example. New content should be in the format: `text' + `pattern' + `label'; `text' + `pattern' + `label'. All the text, patterns and labels are already given as input, if there is no corresponding pattern, just use '' to indicate empty.''\}

\item \{``role'': ``user'', ``content'': ``Each separated text must only have a single label, but may contain several patterns. Each label or pattern must appear at least once in the completion. The patterns can be composed with AND (+) or OR (|) operators.''\}

\item \{``role'': ``user'', ``content'': ``Conversation: Great customer service, reasonable prices, and a chill atmosphere. Pattern: [`(customer)+*+[service]', `(pay)|(sale)', `(environment)'] Label: price, service, environment''\}

\item \{``role": ``assistant", ``content": `` `Great customer service, ' + `(customer)+*+[service]' + `service'; `reasonable prices, ' + `(pay)|(sale)' + `price'; `and a chill atmosphere.' + `(environment)' + `environment' ''\}

\item \{``role'': ``user'', ``content'': ``Conversation: \{\textbf{text}\} Pattern: \{\textbf{pattern}\} Label: \{\textbf{label}\}''\}

\end{itemize}

\subsubsection{GPT-4o Candidate Phrases Generator}
As we are generating counterfactuals that keeps neuro-symbolic patterns, the first step of this task is to generate candidate phrases that keep the pattern but variate semantically, which make up crucial branches of generated counterfactual variations. For this part, we call GPT-4o through the API provided by OpenAI, set the temperature parameter to 0 and restrict the maximum token number to 256. The prompt used is shown below:
\begin{itemize}
\item \{``role'': ``system'', ``content'':``The assistant will create a list of phrases that match the given domain specific language based on the given definition.''\}

\item \{``role'': ``user'', ``content'': ``For the following text and pattern, generate as many diverse example phrases that match the given pattern and can be part of the given target label. Try to not use the word \{\textbf{label}\} or \{\textbf{target\_label}\} in the phrases you generate. Separated your answer by a comma''\}
                
\item \{``role'': ``user'', ``content'': ``text: \{\textbf{matched\_phrase}\}, pattern: \{\textbf{pattern}\}, current label: \{\textbf{label}\} target label: \{\textbf{target\_label}\}''\}

\item \{``role'': ``user'', ``content'': ``The word `\{\textbf{match}\}` is a soft match, you can only use \{\textbf{soft-match\_words}\} as its synonyms to replace it. You can not use other words for \{\textbf{match}\}''\}
\end{itemize}
\subsubsection{GPT-4o Counterfactual Generator}
The GPT-4o generator will finish the second step of counterfactual generation, making use of candidate phrases generated in the first step and combining these semantic pieces into reasonable sentences. We set the temperature parameter to 0 and restrict the maximum token number to 256. The prompt used is shown below:
\begin{itemize}
\item \{``role'': ``system'', ``content'': ``The assistant will generate a counterfactual example close to the original sentence that contains one of the given phrases.''\}

\item \{``role'': ``user'', ``content'': ``Your task is to change the given sentence from the current label to the target.

For example: `Find me a train ticket next monday to new york city' with original label ``transport'' would be turned to `Play me a song called New York City by Taylor Swift' with a label ``audio''.

You can use the following phrases to help you generate the counterfactuals. Please make the sentence about \{\textbf{target\_label}\}. Make sure that the new sentence is not about \{\textbf{label}\}.
You must use one of the following phrases without rewording it in the new sentence: \{\textbf{generated\_phrases}\}''\}

\item \{``role'': ``user'', ``content'': ``You must follow three criteria:

criteria 1: the phrase should change the label from \{\textbf{label}\} to \{\textbf{target\_label}\} to the highest degree.

criteria 2: the modified sentence can not also be about \{\textbf{label}\} and make sure the word \{\textbf{target\_label}\} is not part of the modified sentence.

criteria 3: the modified sentence should be grammatically correct.''\}

\item \{``role'': ``user'', ``content'': ``If you find that you cannot generate new sentence that fulfill all the requirements above, just response `cannot generate counterfactual' and don't feel bad about this''\}

\item \{``role'': ``user'', ``content'': ``original text:\{\textbf{text}\}, original label:\{\textbf{label}\}, modified label:\{\textbf{target\_label}\}, generated phrases:\{\textbf{generated\_phrases}\}, modified text: ''\}
\end{itemize}

\section{Ablation Study on Counterfactual Filtering Methods}
\label{sec:ablation study}

\begin{table*}[htbp]
\centering
\small
\begin{tabular}{@{}rlllllll@{}}

\toprule
\rowcolor{blue!5}
\multicolumn{1}{l}{}             \textbf{No of Shots}            & \textbf{10}   & \textbf{15}   & \textbf{30}   & \textbf{50}   & \textbf{70}   & \textbf{90}   & \textbf{120}  \\ \midrule
\multirow{2}{*}{\begin{tabular}[c]{@{}r@{}} No Filters \\ SD\end{tabular} }                      &  0.10    &  0.12    &  0.15    &   0.23   &  0.23    &   0.21   &  0.21    \\ \cmidrule(l){2-8} 
                                             &  0.03    &  0.04    &  0.05    &   0.04   &  0.04    &    0.03  &  0.03    \\ \midrule

\multirow{2}{*}{\begin{tabular}[c]{@{}r@{}}Herustic Filter \\ SD\end{tabular} }             &  0.15    &   0.17   &   0.19   &  0.28    &  0.27    &  0.28   &  0.28    \\ \cmidrule(l){2-8} 
                                             &  0.08    &  0.1   &  0.1    &  0.07     &  0.09    &    0.1  &   0.1   \\ \midrule
\multirow{2}{*}{\begin{tabular}[c]{@{}r@{}}Herustic + Symbolic Filters \\ SD\end{tabular}} & 0.12     &  0.13    &   0.13   &   0.17   &   0.16   &   0.18   &   0.20   \\ \cmidrule(l){2-8} 
                                             &   0.04   &  0.03    &    0.01        &   0.02  & 0.03 &  0.02    &    0.01  \\ \midrule
\multirow{2}{*}{\begin{tabular}[c]{@{}r@{}}Herustic + LLM Discriminator \\ SD\end{tabular}} & 0.17     &  0.21    &   0.23   &   0.34   &   0.42  &   0.45   &   0.49   \\ \cmidrule(l){2-8} 
                                             &   0.08   &  0.04    &    0.09        &   0.07  & 0.02 &  0.02    &    0.05  \\ \midrule
\multirow{2}{*}{\begin{tabular}[c]{@{}r@{}}Herustic + Symbolic + LLM Discriminator \\ SD\end{tabular}} & \textbf{0.38} & \textbf{0.39} & \textbf{0.49} & \textbf{0.47} & \textbf{0.51} & \textbf{0.53} & \textbf{0.50} \\ \cmidrule(l){2-8} 
                                             & 0.04 & 0.08 & 0.06 & 0.04 & 0.05 & 0.05 & 0.04 
                                             \\   \bottomrule
\end{tabular}
\caption{Average F1-score and SD from an ablation study with the YELP dataset on BERT model}
\label{tab:yelp_ablation}
\end{table*}

We performed an ablation study to investigate the impact of the different components in our filtering pipeline. We follow the same approach as \S~\ref{sec:active_learning_eval} where each condition is run with different seeds 8 times. For each condition, we report an average F1 score and the standard deviation (SD) in Table~\ref{tab:yelp_ablation}. Our approach involves generating counterexamples with a fine-tuned GPT-4o model and applying all three filters defined in \S~\ref{sec: filter} before using the data for active learning.

In this study, we investigate the impact of different configurations by varying the filtering mechanisms used with the generator model.

The ablation study is conducted using the YELP dataset with a BERT model for the downstream active learning tasks. The configurations tested include:
\begin{itemize}
    \item No Filters: Counterexamples generated without any filters applied
    \item Heuristic Filter: Applying only the heuristic filter
    \item Heuristic + Symbolic Filters: Applying both heuristic and symbolic filters
    \item All Filters: Applying all three filters defined in \S~\ref{sec: filter}
\end{itemize}

The results indicate that the use of all filters significantly improves the performance of the trained model~(See Table~\ref{tab:yelp_ablation}). The average F1-score with all filters applied reaches 0.51 for 70 shots and peaks at 0.53 for 90 shots, demonstrating a 2X improvement over the baseline with no filters (F1-score of 0.23 for 70 shots). Using a pairwise t-test, we find that this is statistically significant (p<0.0001), underscoring the value of carefully filtering LLM-generated counterfactuals to produce usable data for model training.

\begin{table*}[t]
\centering
\scriptsize
\setlength{\tabcolsep}{8pt}
\renewcommand{\arraystretch}{1.0}

\vspace{.3em}

% YELP - GPT-4o
\textbf{Counterfactual Generated using Llama3.3} \\

\resizebox{\textwidth}{!}{\begin{tabular}{rlllllll}
\toprule
\rowcolor{blue!5}
\textbf{Method} & 10 & 15 & 30 & 50 & 70 & 90 & 120 \\
\midrule
YELP & \fscorestar{.31}{.05}{} & \fscorestar{.32}{.06}{} & \fscorestar{.34}{.08}{} & \fscorestar{.44}{.05}{} & \fscorestar{.51}{.08}{} & \fscorestar{.53}{.04}{} & \fscorestar{.64}{.04}{} \\
MASSIVE & \fscorestar{.28}{.08}{} & \fscorestar{.36}{.06}{} & \fscorestar{.40}{.02}{} & \fscorestar{.54}{.07}{} & \fscorestar{.58}{.06}{} & \fscorestar{.60}{.03}{} & \fscorestar{.66}{.03}{}  \\
EMOTIONS & \fscorestar{.21}{.09}{} & \fscorestar{.24}{.05}{} & \fscorestar{.32}{.1}{} & \fscorestar{.34}{.07}{} & \fscorestar{.39}{.1}{} & \fscorestar{.47}{.06}{} & \fscorestar{.51}{.08}{} \\
\bottomrule
\end{tabular}}

\caption{Macro F1-scores for the counterfactual condition using Llama3.3 as the counterfactual generator model for BERT, evaluated across three datasets (YELP, MASSIVE, EMOTIONS) at varying annotation shot counts. }
\label{tab:llama-scores}
\end{table*}

Surprisingly, we found that incorporating the symbolic filter without the LLM discriminator decreases the performance of downstream training. Further analysis of the included examples revealed that some generated sentences included the original sentence with additional parts that corresponded to the target label. While the LLM discriminator would filter these out, without its use in the pipeline, these generated counterfactuals are mistakenly treated as negative examples, when technically they are just multi-labeled positive examples. However, we observe a substantial improvement in performance when the symbolic filter is used in conjunction with the LLM discriminator, as opposed to using the LLM discriminator alone. This demonstrates the effectiveness of combining both methods to enhance the quality and accuracy of the generated counterfactuals.

The ablation study highlights the crucial role of the filtering pipeline. By systematically evaluating the impact of each component, we demonstrate that the integration of heuristic, symbolic filters, and the LLM discriminator leads to significant improvements in the downstream active learning task. This validates our hypothesis that filtering LLM-generated data is essential in determining usable and useful data for achieving higher performance and reliability in model training.

\section{Experiment using Open-weight Counterfactual Generator}
\label{sec:llama-experiment}

We evaluate the effectiveness of our counterfactual generation approach using an open-weight model Llama3.3 model as the generator with BERT as the classifier model across three datasets (YELP, MASSIVE, EMOTIONS), under increasing annotation shot counts. Our findings as seen in Table~\ref{tab:llama-scores} show that performance using the Llama3.3 model is comparable across all datasets, showing the viability of our method beyond proprietary models.

\end{document}